\documentclass[11pt]{article}

\usepackage[final]{acl}

\usepackage{times}
\usepackage{latexsym}

\usepackage[T1]{fontenc}

\usepackage[utf8]{inputenc}

\usepackage{microtype}

\usepackage{inconsolata}

\usepackage{acronym}

\usepackage{hyperref}

\usepackage{dirtytalk} 

\usepackage{listings}
\usepackage{xcolor}
\usepackage{moreverb}
\usepackage{alltt}
\usepackage{fancyvrb}
\usepackage{tcolorbox}
\tcbuselibrary{listingsutf8}

\newcommand{\promptlisting}[3]{
  \lstinputlisting[
    basicstyle=\scriptsize\ttfamily,
    breaklines=true,
    breakatwhitespace=true,
    breakindent=0pt,
    frame=single,
    numbers=none,
    tabsize=2,
    showstringspaces=false,
    backgroundcolor=\color{gray!10},
    caption=#2,
    label=#3
  ]{#1}
}
\newcommand{\promptlistingsmall}[3]{
  \lstinputlisting[
    basicstyle=\small\ttfamily,
    breaklines=true,
    breakatwhitespace=true,
    breakindent=0pt,
    frame=single,
    numbers=none,
    tabsize=2,
    showstringspaces=false,
    backgroundcolor=\color{gray!10},
    caption=#2,
    label=#3
  ]{#1}
}
\usepackage{amsmath}
\usepackage{amsfonts}

\DeclareMathOperator*{\argmin}{arg\,min}

\acrodef{NLP}{Natural Language Processing}
\acrodef{ML}{Machine Learning}
\acrodef{LLM}{Large Language Model}
\acrodefplural{LLM}{Large Language Models}
\acrodef{ORM}{Outcome Reward Model}
\acrodef{PRM}{Process Reward Model}
\acrodefplural{PRM}{Process Reward Models}
\acrodef{MC}{Monte Carlo}
\acrodef{BoN}{Best-of-N}
\acrodef{CoT}{Chain-of-Thought}
\acrodefplural{CoT}{Chains-of-Thought}
\acrodef{MoE}{Mixture-of-Experts}
\acrodef{NLI}{Natural Language Inference}

\usepackage{graphicx}

\usepackage{amssymb}
\usepackage{pifont}
\newcommand{\cmark}{\ding{51}}%
\newcommand{\xmark}{\ding{55}}%

\title{BoN Appetit Team
at LeWiDi-2025: \\ Best-of-N Test-time Scaling Can Not Stomach Annotation\\
Disagreements (Yet)}

\author{
  Tomas Ruiz{\normalfont \textsuperscript{1,2}} \qquad 
  Siyao Peng{\normalfont \textsuperscript{1,3}} \qquad
  Barbara Plank{\normalfont \textsuperscript{1,3}} \qquad
  Carsten Schwemmer{\normalfont \textsuperscript{1,2}}\\
  \textsuperscript{1}Ludwig Maximilian University of Munich, Germany \\
  \textsuperscript{2}Computational Social Sciences \qquad
  \textsuperscript{3}MaiNLP \& MCML \\
  \texttt{\{t.ruiz,siyao.peng,b.plank,carsten.schwemmer\}@lmu.de}
}

\begin{document}
\maketitle

\begin{abstract}
Test-time scaling is a family of techniques to improve LLM outputs at inference time by performing extra computation.
To the best of our knowledge, test-time scaling has been limited to domains with verifiably correct answers, like mathematics and coding.
We transfer test-time scaling to the LeWiDi-2025 tasks to evaluate annotation disagreements.
We experiment with three test-time scaling methods: two benchmark algorithms (Model Averaging and Majority Voting), and a \ac{BoN} sampling method.
The two benchmark methods improve LLM performance consistently on the LeWiDi tasks, but the \ac{BoN} method does not.
Our experiments suggest that the \ac{BoN} method does not currently transfer from mathematics to 
LeWiDi tasks, and we analyze potential reasons for this gap.
\end{abstract}

\section{Introduction}
Supervised learning typically assumes a single fixed label per example.
However, prior work documents substantial interpretative variability in human annotations, with annotators often disagreeing on labels~\cite{https://doi.org/10.17185/duepublico/42132,warner-hirschberg-2012-detecting,baan-etal-2022-stop}, especially for subjective \ac{NLP} tasks~\cite{ovesdotter-alm-2011-subjective}.
~\citet{plank-2022-problem} and \citet{Cabitza_2023} argue that this variability is informative rather than problematic and~\citet{rottger-etal-2022-two} suggests that variability should be explicitly integrated into the annotation processes.

The shared task \textbf{Learning With Disagreement} (LeWiDi) 2025~\cite{LeWiDi2025} tackles this opportunity and provides four datasets with annotator-level metadata and label variation. We document the datasets in detail in \autoref{sec:datasets}. The datasets support two different tasks: (1) Perspecivist task: Predicting the label of each individual annotator. (2) Soft-label task: Predicting the distribution of human annotations for a single problem instance.
This distribution is known as a \textbf{soft-label}, or a \textit{human judgement distribution}~\cite{nie-etal-2020-learn}.

\begin{figure}[t]
  \centering
  \includegraphics[width=\columnwidth]{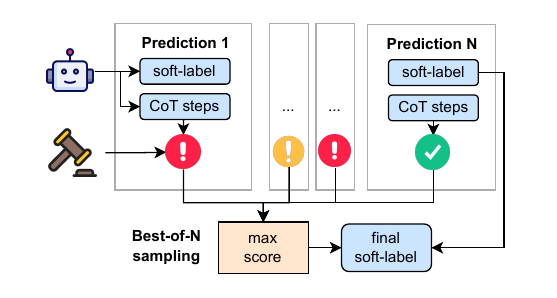}
  \caption{\textbf{Best-of-N sampling with step-wise scores}.
  For each problem a reasoning \acs{LLM} generates $N$ soft-labels and \acp{CoT}.
  Next, an LLM-as-a-judge scores each step in the \ac{CoT} for correctness, and \ac{BoN} selects the soft-label with the highest total score.
  \textit{Takeaway}: Sampling multiple times increases the chances for a good prediction.}
  \label{fig:best-of-n-diagram}
\end{figure}

In the previous iteration of the LeWiDi shared task~\cite{leonardelli-etal-2023-semeval}, 
many teams trained encoder-based models like BERT~\cite{devlin-etal-2019-bert} directly on soft-labels.
However, the innovations in generalist \acp{LLM} and the rise of ``reasoning'' capabilities~\cite{wei2023chainofthoughtpromptingelicitsreasoning,openai2025o3-o4-mini,yang2025qwen3technicalreport,deepseekai2025deepseekr1incentivizingreasoningcapability} motivated us to answer the following question:

\say{\textit{Can reasoning \acp{LLM} handle interpretative variability and annotation disagreement effectively at inference time?}}

To answer this question, we turn to \textbf{test-time scaling} methods, like \ac{BoN} sampling, which improve the \acp{LLM} performance by spending more compute per problem~\cite{cobbe2021trainingverifierssolvemath,shen-etal-2021-generate-rank}.
These methods have been very successful in mathematics and coding, but have not been applied yet to \ac{NLP} tasks with annotation disagreement, as far as we know. 
In this paper, we take established test-time scaling methods and apply them to the LeWiDi tasks.

Our \textbf{contributions} are:
\begin{itemize}
  \item A metric named prediction diversity, used to analyze the performance of test-time scaling methods on soft-label tasks. We show that it tracks problem difficulty on the LeWiDi tasks.
  \item We show that Model Averaging and Majority Voting consistently improve \acp{LLM} performance across all LeWiDi datasets.
  \item Finally, we show that \ac{BoN} sampling with step-wise scores \textit{does not work well} on the LeWiDi tasks, and analyze potential causes.
\end{itemize}

\section{Related Work}

\subsection{Learning Interpretative Variability}
Modeling the diverse perspectives that human annotators have on the same problems is important to prevent minority voices from being ignored~\cite{leonardelli-etal-2021-agreeing}.
Prior work on modeling annotator disagreement has explored various techniques, such as using separate model heads for each annotator~\cite{davani-etal-2022-dealing}, learning specific representations for annotators~\cite{mokhberian-etal-2024-capturing}, separating stable opinions from annotation mistakes~\citep{gordon2021disagreement,weber-genzel-etal-2024-varierr}, and using soft-labels to aid learning~\citep{fornaciari-etal-2021-beyond, uma_case_2020}.
To evaluate models on soft-labels, ~\citet{rizzi-etal-2024-soft} propose using the Manhattan or Euclidean distance rather than the Cross-Entropy loss.
In terms of quantifying the diversity of soft-labels, \citet{singh-etal-2024-learning} proposed the Jensen-Shannon Divergence in the context of ensemble classification.

\subsection{Test-Time Scaling}
Test-time scaling methods improve the performance of \acp{LLM} by spending more compute per problem instance.
One approach is to refine an initial response iteratively with self-feedback \citep{madaan2023selfrefineiterativerefinementselffeedback},
or improve the response by following a set of rules (a \textit{constitution}, \citealt{bai2022constitutionalaiharmlessnessai}).
Another common test-time scaling approach is \textbf{\acl{BoN} sampling}, where multiple solutions are sampled in parallel, and a verifier model scores or ranks the solutions to select the best one~\cite{cobbe2021trainingverifierssolvemath,shen-etal-2021-generate-rank}. 
The scores are computed based solely on the outcome (correct or incorrect) of the task (\textit{\acl{ORM})}.
But the scores can also be computed for the correctness of individual reasoning steps used to arrive at the answer (\textit{\acl{PRM}}).
~\citet{lightman2023letsverifystepstep} showed that scoring individual steps in a \acf{CoT} for correctness, and discarding \acp{CoT} with faulty steps improves the performance of \acp{LLM} on the MATH dataset~\cite{hendrycks2021measuringmathematicalproblemsolving}.
In their work, the scoring annotations were provided by humans. Follow-up work replaced the human scores with \textit{automated scoring}, using either \ac{MC} sampling~\cite{wang-etal-2024-math} or an LLM judge (\textit{LLM-as-a-judge}, \citealp{zheng2023judgingllmasajudgemtbenchchatbot}).
Research by~\citet{zhang-etal-2025-lessons} and~\citet{zheng-etal-2025-processbench} showed that using \acp{LLM} to provide these scores generalizes well and is competitive with training a custom model.

\noindent A different approach that leverages diversity plus selection is \textit{\ac{MoE}}: multiple parallel expert subnetworks, with a gate that selects a few experts per input. Both \ac{MoE} and test-time scaling are independent approaches that can be combined during model evaluation, \emph{e.g.}\ as did~\citet{comanici2025gemini25pushingfrontier} for the SWE-Bench~\cite{jimenez2024swebench}.

\section{Method}
\paragraph{Nomenclature:} A dataset is a collection of problem instances (\textit{problem} in short).
We sample a reasoning \ac{LLM} $N$ times to solve a problem. Each \textit{sample} contains a \textit{prediction} and a \textit{\ac{CoT}}.
A prediction could be text, a soft-label or a list of integers (perspectivist task).

Our test-time scaling method is not novel but rather a combination of methods already established in the literature.
Our innovation is to apply it to a new domain: the LeWiDi-2025 tasks.
We refer to the method as \textit{\ac{BoN} sampling with step-wise scores}, or just \textit{\ac{BoN} sampling} in short.
The method, shown in \autoref{fig:best-of-n-diagram}, consists of three steps: (1) A reasoning \ac{LLM} generates $N$ samples for a problem.
(2) A judge \ac{LLM} scores each \ac{CoT}-step in each sample for correctness.
(3) We choose the sample with the best score for the final prediction.
We explain all method details in \autoref{sec:llm-setup} and \autoref{sec:bon-sampling-with-scoring}.
~\autoref{tab:methods} is an overview of all methods we run experimentally, which include different baselines and benchmarks.

\begin{table}[h]
  \centering
  \begin{tabular}{lcc}
    \textbf{Method} & \textbf{Samples} & \textbf{Use \ac{CoT}} \\
    \hline
    Most Frequent & - & - \\
    Simple Sampling & 1 & \xmark \\
    \textit{Model Averaging} & $N$ & \xmark \\
    \textit{Majority Voting} & $N$ & \xmark \\
    BoN Oracle & $N$ & \xmark \\
    \ac{BoN} + SWS & $N$ & \cmark \\
  \end{tabular}
  \caption{\textbf{Methods Overview.}
  The first two methods are baselines (\autoref{sec:baselines}).
  The next two methods are our own benchmarks (\autoref{sec:test-time-scaling-benchmarks}).
  The \ac{BoN} Oracle (\autoref{sec:upper-bound-on-performance}) is a performance upper bound on our proposed method \ac{BoN} + SWS (Step-Wise Scores, \autoref{sec:bon-sampling-with-scoring}).
  The models we submitted to the shared task are in \textit{italic}.}
  \label{tab:methods}
\end{table}

\subsection{LLM Setup}
\label{sec:llm-setup}
\paragraph{Prompts} We prompt a reasoning \ac{LLM} to solve the soft-label and perspectivist tasks directly. 
For example, in the soft-label task, we present the dataset (\emph{e.g.}\ sarcasm detection), and instruct the model to predict the human soft-labels (snippet in \autoref{lst:csc-soft-label-snippet}, full prompt in \autoref{lst:csc-soft-label}).

\promptlistingsmall{listings/CSC_t60_snippet.txt}{Prompt Snippet (Soft-label Task)}{lst:csc-soft-label-snippet}

\noindent In the perspectivist task, we instead instruct the model to predict the label for each annotator (snippet in \autoref{lst:csc-perspectivist-snippet}, full prompt in \autoref{lst:csc-perspectivist}).

\promptlistingsmall{listings/CSC_t63_snippet.txt}{Prompt Snippet (Perspectivist Task)}{lst:csc-perspectivist-snippet}

\noindent We include a prompt section that explicitly instructs the model to reason about the \textit{diverse perspectives and interpretations} that annotators could have.
This improved performance by a small, but statistically significant margin, so we included it in all later experiments (see \autoref{sec:prompt-ablations}).

\subsection{\ac{BoN} Sampling with Step-Wise Scores}
\label{sec:bon-sampling-with-scoring}
In \ac{BoN} sampling we score each of the $N$ model samples and select the best for the final prediction.
The score for a sample depends on the correctness of each step in its \ac{CoT}.
Our \ac{BoN} sampling method borrows heavily from~\citet{lightman2023letsverifystepstep},
so in \autoref{tab:method-comparison-lightman-et-al} we summarize the differences and similarities between both.

\begin{table}[h]
  \centering
  \begin{tabular}{lll}
  & \textbf{Lightman et al.} & \textbf{Ours} \\
  \hline
  Domain & Math & LeWiDi Tasks \\
  Model & GPT-4 & Qwen3-32B \\
  Scorer & Human(s) & LLM-as-a-judge \\
  Reduction & Product & Mean \\
  \hline
  Sampling & \multicolumn{2}{c}{Best-of-N} \\
  Scores & \multicolumn{2}{c}{bad=0, okay=0, good=1} \\
  \end{tabular}
  \caption{\textbf{Comparison to \citet{lightman2023letsverifystepstep}.} First 4 rows are differences, last 2 rows are similarities.}
  \label{tab:method-comparison-lightman-et-al}
\end{table}

\noindent First, we split the \ac{CoT} into logical steps (details in \autoref{sec:splitting-cot-into-steps}), and then score each step as either ``great'', ``okay'', or ``bad'' in line with \citet{lightman2023letsverifystepstep}.
They used human annotations to train a scoring model, but we use an \textit{LLM-as-a-judge} instead to provide the scores directly, as suggested by~\citet{zheng-etal-2025-processbench}.
The prompt for the LLM-as-a-judge is based on their scoring instructions 
(snippet in \autoref{lst:judge-prompt-snippet}, full prompt in \autoref{lst:judge-prompt}).

\promptlistingsmall{listings/judge-prompt_snippet.txt}{Prompt Snippet (LLM-as-a-judge)}{lst:judge-prompt-snippet}

\noindent We follow \citet{lightman2023letsverifystepstep} to convert the three scores to numbers 
(bad=0, okay=0, good=1), and average all the step-wise scores to compute a \textit{prediction-level score}.
\citet{lightman2023letsverifystepstep} used a product reduction, but in our experiments, mean reduction outperformed product.
This step-wise scoring is repeated for $N=10$ model samples and their corresponding \acp{CoT}.
We select the one with the highest prediction-level score as the final prediction.

\paragraph{Models}
Previous research in hate speech detection and \ac{NLI} showed that \textit{explanations} are useful to judge the plausibility and correctness of annotations and model predictions~\citep{mathew2022hatexplainbenchmarkdatasetexplainable,jiang-etal-2023-ecologically,weber-genzel-etal-2024-varierr}.
We hypothesize that reasoning about annotator disagreements also requires a \textit{deliberative} and \textit{explanatory} approach that considers multiple interpretations and weights their likelihoods.
Therefore, we use a reasoning \ac{LLM} for our experiments (Qwen3-32B,~\citealt{yang2025qwen3technicalreport}).
For LLM-as-a-judge we use a model from a different family besides Qwen3 (DeepSeek-R1-0528-Qwen3-8B,~\citealt{deepseekai2025deepseekr1incentivizingreasoningcapability}). Sampling parameters are detailed in \autoref{sec:sampling-parameters}.

\subsection{Upper Bound on Performance}
\label{sec:upper-bound-on-performance}
We determine the upper bound on performance of \textit{any} \ac{BoN} sampling method by computing a so-called \textbf{\ac{BoN} oracle}.
The \ac{BoN} oracle is a hypothetical model that always selects the best prediction among $N$ predictions (in our case, lowest distance).
We compute the oracle in the \textit{training} set by choosing the soft-label among the $N$ predictions with the lowest distance to the human soft-label.
However, for a dataset with unknown human soft-labels, we cannot compute the oracle.
The oracle is an analytical tool to determine the \ac{BoN} performance ceiling, rather than an algorithm to use in practice.
The oracle soft-label $p_o$ for a set of predictions $P$ (size $N$) is defined as:
\begin{equation}
  p_o = \argmin_{p \in P} W(p, p_h)
\end{equation}
where $p_h$ is the human soft-label. We compute it for both the soft-label and perspectivist tasks.

\subsection{Baselines}
\label{sec:baselines}
The LeWiDi 2025 shared task proposed the \textbf{Most Frequent Baseline}. 
In the soft-label task, this is the mean label value for each label across all training problems.
In the perspectivist task, it is the most frequent label for each individual annotator.
Our own basic baseline is the performance of the \ac{LLM} without any test-time scaling, \emph{i.e.}\ with a single sample per problem ($N=1$).
We call this \textbf{Simple Sampling}.

\subsection{Test-Time Scaling Benchmarks}
\label{sec:test-time-scaling-benchmarks}
\ac{BoN} sampling uses a lot more compute per problem than Simple Sampling ($N$ times more).
To benchmark \ac{BoN} sampling fairly, we compare it with two test-time scaling algorithms that also create a single prediction out of $N$ predictions.

\paragraph{Soft-label task} We benchmark against \textit{Model Averaging}, where all $N$ soft-labels $p^{n}$ are averaged into a single soft-label $\bar{p}$. 
The resulting soft-label $\bar{p}$ is a valid probability distribution. 
Each entry $i$ of $\bar{p}$ is defined as:
\begin{equation}
  \bar{p_i} = \frac{1}{N} \sum_{n=1}^N p^{n}_i
\end{equation}

\paragraph{Perspectivist task} We benchmark against \textit{Majority Voting}, where we sample the model $N$ times per problem, each prediction resulting in a label per annotator, and then select the most frequent label (within the $N$ predictions) for each annotator.

\subsection{LeWiDi-2025 Datasets}
\label{sec:datasets}
We report the datasets as provided by the LeWiDi-2025 shared task.
All datasets provide some level of annotator-level metadata like gender, age, nationality, education and more.

\paragraph{The Conversational Sarcasm Corpus (CSC):}
The CSC dataset by~\citet{jang-frassinelli-2024-generalizable} is a dataset for sarcasm detection with 7,036 entries (5,628 train, 704 dev, 704 test).
Each entry consists of a context+response pair, where the reponse is rated for sarcasm on a 6-point Likert scale, by either 4 or 6 annotators.

\paragraph{The MultiPico dataset (MP):}
The MP by~\citet{casola-etal-2024-multipico} is a dataset for irony detection with 18,778 entries (12,017 train, 3,005 dev, 3,756 test).
Each entry consists of a post-reply pair from Twitter and Reddit, and the reply's irony is rated as either ironic (1) or not ironic (0) by between 2 and 21 annotators.

\paragraph{The Paraphrase Detection dataset (PAR):}
The Paraphrase is a dataset by the MaiNLP lab\footnote{https://mainlp.github.io/} for paraphrasing detection with 500 entries (400 train, 50 dev, 50 test). 
Each entry has two questions from Quora Question Pairs (QQP), and annotators rate how strongly the questions are paraphrases of one another from -5 to 5.
Each entry is rated by 4 annotators.

\paragraph{The VariErr NLI dataset (VEN):}
VariErrNLI by~\citet{weber-genzel-etal-2024-varierr} is a dataset for \ac{NLI} with 500 samples (400 train, 50 dev, 50 test). Annotators can assign any and multiple of the \ac{NLI} categories (entailment, contradiction, neutral) for each entry. Each entry is annotated by 4 annotators.

\subsection{Metrics}

\paragraph{Soft-label Task}
As suggested by the LeWiDi task, we report \textit{Manhattan Distance} for the MP and VEN datasets, and \textit{Wasserstein Distance} for the CSC and PAR datasets.
Both distances are exactly equivalent when applied to binary datasets~\citep{rizzi-etal-2024-soft}.
The Wasserstein distance measures the minimum ``work'' needed to transform one probability distribution into another, where ``work'' equals the amount of mass moved times the distance.

\paragraph{Perspectivist Task}
For the perspectivist task, we report \textit{Error Rate} (1 - accuracy) for the MP and VEN datasets, and \textit{Absolute Distance} for the CSC and PAR datasets.
We divide the Absolute Distance by the range of the Likert scale, in line with the LeWiDi organizers.
Both metrics are exactly equivalent when applied to binary datasets.

\paragraph{Prediction Diversity}
\ac{BoN} sampling requires a diverse set of predictions for each problem.
Otherwise, if all predictions were the same (or very similar), it would not matter which one is selected, and \ac{BoN} sampling would provide no improvement over Simple Sampling. 
Therefore, we quantify the variability of the soft-labels across the $N$ predictions for each problem, and call this the \textit{prediction diversity}.

We implement this as the average pair-wise distance between all $N$ soft-labels for a single problem.
For the LeWiDi datasets, we use the Wasserstein distance because it can capture distances in Likert scales.
We do not compare soft-labels to themselves, because the distance is 0.
This is why we divide by $N(N-1)$ rather than by $N^2$. The formula for diversity $D$ is:
\begin{equation}
  \label{eq:prediction_diversity}
  D(P) = \frac{1}{N(N-1)} \sum_{i=1}^N \sum_{j \neq i}^N W(p^i, p^j)
\end{equation}
where $P$ is the set of $N$ soft-labels, and $W(p^i, p^j)$ is the Wasserstein distance between the soft-labels $p^i$ and $p^j$.
Note that measuring the diversity of a set of predictions $P$ is different from measuring the spread of a single soft-label (\emph{i.e.}\ measuring the entropy of the soft-label).

\paragraph{Problem Difficulty}
Classically, problem difficulty is measured as the percentage of correct answers over $N$ attempts. 
For soft-label tasks, it can instead be defined as the distance between predictions and human soft-label across $N$ attempts.
In the LeWiDi task, we posit a relationship between prediction diversity and problem difficulty: 
low diversity arises when the model perceives no ambiguity (the problem is easy or only one interpretation is considered), while high diversity arises when multiple plausible interpretations exist and the model's $N$ predictions vary.
Since both prediction diversity and distances (Wasserstein, Manhattan) are computable, their correlation is empirically measurable.

\section{Results}
In \autoref{tab:results-test} we summarize the results on the test, taken from the LeWiDi leaderboard\footnote{LeWiDi leaderboard: \url{https://le-wi-di.github.io/}}, since we have no access to the test set ground truth.
In \autoref{tab:results-train} we present a performance overview of all methods for the LeWiDi datasets.
We did not train on the train set, so we used it for evaluation.


\begin{table}[h]
  \centering
  \resizebox{0.48\textwidth}{!}{
  \begin{tabular}{lcc}
    & \multicolumn{2}{c}{\textbf{Task}} \\
    \cline{2-3}
    \textbf{Dataset} & Soft-label ($\downarrow$) & Perspectivist ($\downarrow$) \\
    \hline
    CSC & 0.928 & 0.231 \\
    MP & 0.466 & 0.414 \\
    PAR & 1.797 & 0.228 \\
    VEN & 0.356 & 0.272 \\
    \hline
    Avg. Rank & 5th & 6th\\
    Out of & 15 & 11 \\
    \hline
    \textbf{Method} & Model Averaging & Majority Voting
  \end{tabular}
  }
  \caption{Results on the \textbf{test} set of the LeWiDi datasets (lower is better). Values are from the LeWiDi leaderboard. We submitted our best performing methods, Model Averaging and Majority Voting.}
  \label{tab:results-test}
\end{table}

\begin{table*}[h]
  \centering
  \begin{tabular}{lcccc|cccc}
    & \multicolumn{4}{c|}{\textbf{Soft-label Task}} & \multicolumn{4}{c}{\textbf{Perspectivist Task}}\\
    \cline{2-5} \cline{6-9}
    \textbf{Method} & CSC & Par & MP & VEN & CSC & Par & MP & VEN \\
    \hline
    Most Frequent Baseline & 1.14 & 2.89 & 0.26 & 0.27 & \textbf{0.21} & 0.36 & \textbf{0.30} & 0.33 \\
    Simple Sampling & 1.00 & 1.96 & 0.26 & 0.22 & 0.24 & \textbf{0.25} & 0.43 & 0.32 \\
    \textit{Model Averaging} & \textbf{0.91} & \textbf{1.78} & \textbf{0.24} & \textbf{0.20} & - & - & - & - \\
    \textit{Majority Voting} & - & - & - & - & 0.23 & \textbf{0.25} & 0.40 & \textbf{0.30} \\
    BoN Sampling + SWS & 1.01 & 1.93 & 0.26 & 0.22 & 0.24 & \textbf{0.25} & 0.42 & 0.32 \\
    \hline
    BoN Oracle & 0.51 & 1.29 & 0.11 & 0.11 & 0.15 & 0.18 & 0.18 & 0.16 \\
    \hline
    \textbf{Metric ($\downarrow$)} & \multicolumn{2}{c}{Wasserstein} & \multicolumn{2}{c|}{Manhattan} & \multicolumn{2}{c}{Abs. Dist.} & \multicolumn{2}{c}{Error Rate} \\
  \end{tabular}
  \caption{Results on the \textbf{train} set of the LeWiDi datasets.
  In \textbf{bold} is the best performing method by column. 
  \ac{BoN} sampling underperforms the test-time scaling benchmarks, even though the \ac{BoN} oracle suggests a high performance ceiling.
  We submitted to the LeWiDi shared task the Model Averaging (soft-label task) and Majority Voting (perspectivist task) methods, since they perfomed best.}
  \label{tab:results-train}
\end{table*}


\subsection{Test-Time Scaling Benchmarks}
\begin{figure}[h]
  \centering
  \includegraphics[width=\columnwidth]{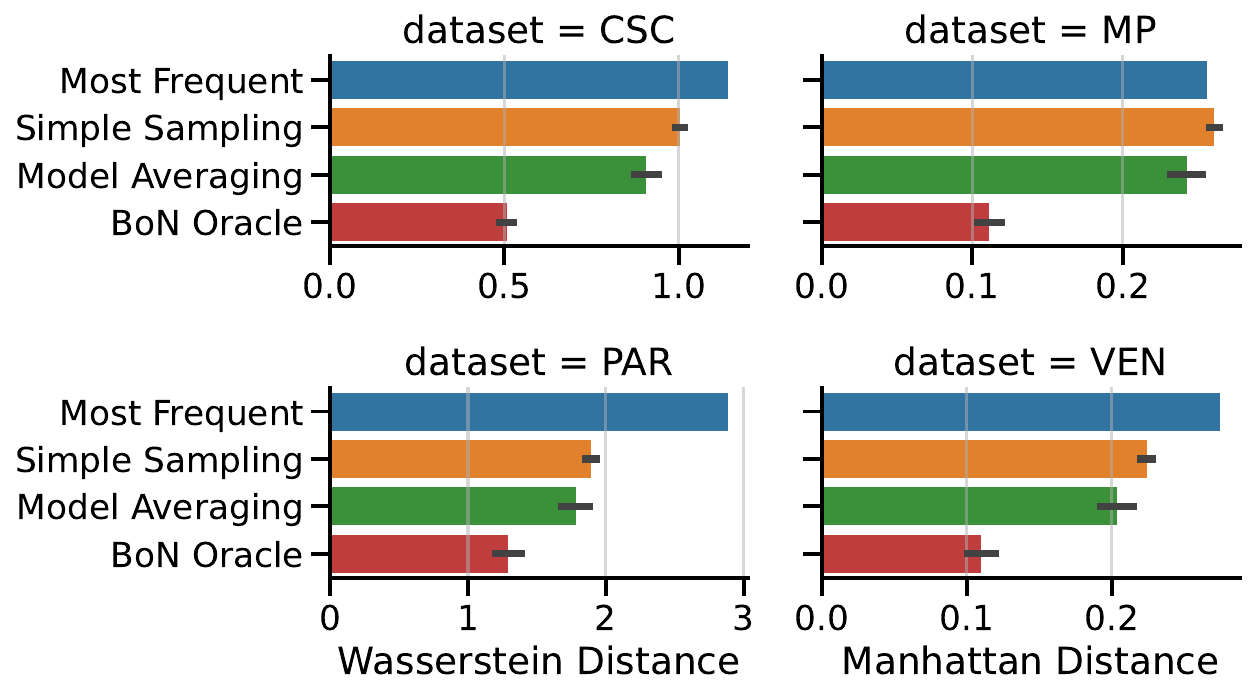}
  \par\vspace{1.25em}\par
  \includegraphics[width=\columnwidth]{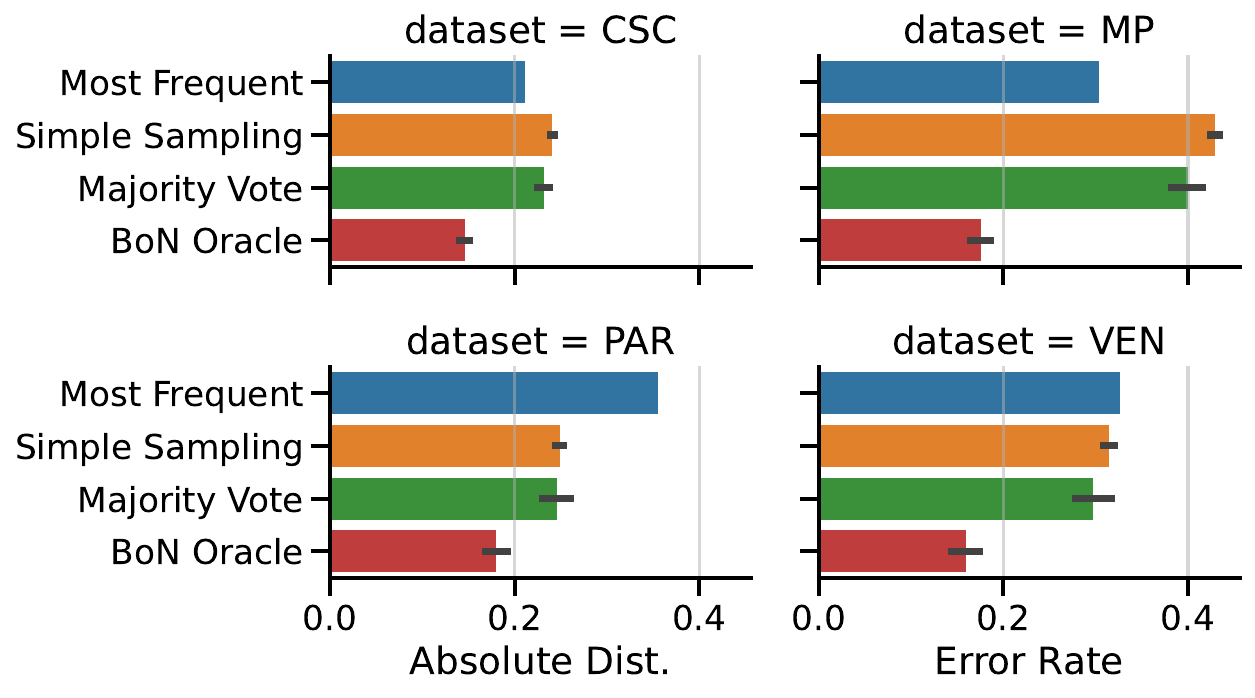}
  \caption{\textbf{Test-time Scaling Benchmarks}.
  Top: soft-label task. Bottom: perspectivist task.
  Distance metric on the x-axis (lower is better). 
  Model Averaging and Majority Voting (green) are consistently better than Simple Sampling (orange) in the soft-label and perspectivist tasks, respectively.}
  \label{fig:soft-label-baselines}
\end{figure}

We first report the performance of all methods except \ac{BoN} sampling.
The orange bar in \autoref{fig:soft-label-baselines} is the performance \textit{Simple Sampling} with Qwen3-32B.
\noindent It outperforms the \textit{Most Frequent Baseline} on 3 out of 4 datasets in the soft-label task, but only in 2 out of 4 datasets in the perspectivist task.
The test-time scaling benchmarks (green) are \textit{Model Averaging} and \textit{Majority Voting}.
Both methods consistently improve performance over Simple Sampling across datasets and tasks.
We discuss the effects of Model Averaging on soft-label entropy in~\autoref{sec:entropy_analysis} and compare it with naive soft-label smoothing.
The \textit{\ac{BoN} oracle} (red) is meant to show the performance ceiling of any \ac{BoN} sampling method. 
Its strong performance indicates that, at least \textit{theoretically}, a good \ac{BoN} sampling method can achieve very good performance on the LeWiDi tasks.

\subsection{Best-of-N Sampling with Step-Wise Scores}
\begin{figure}[!h]
  \centering
  \includegraphics[width=\columnwidth]{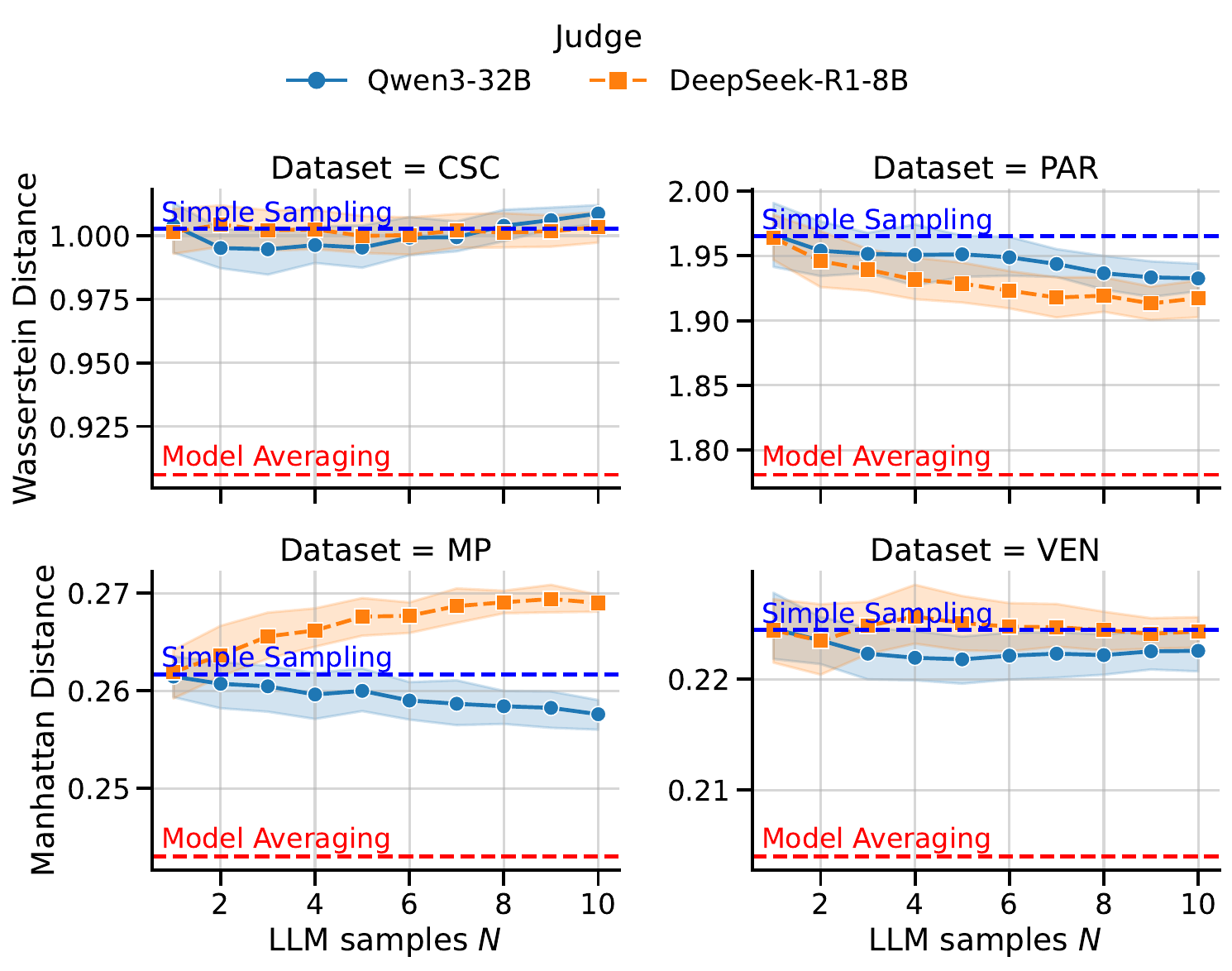}
  \includegraphics[width=\columnwidth]{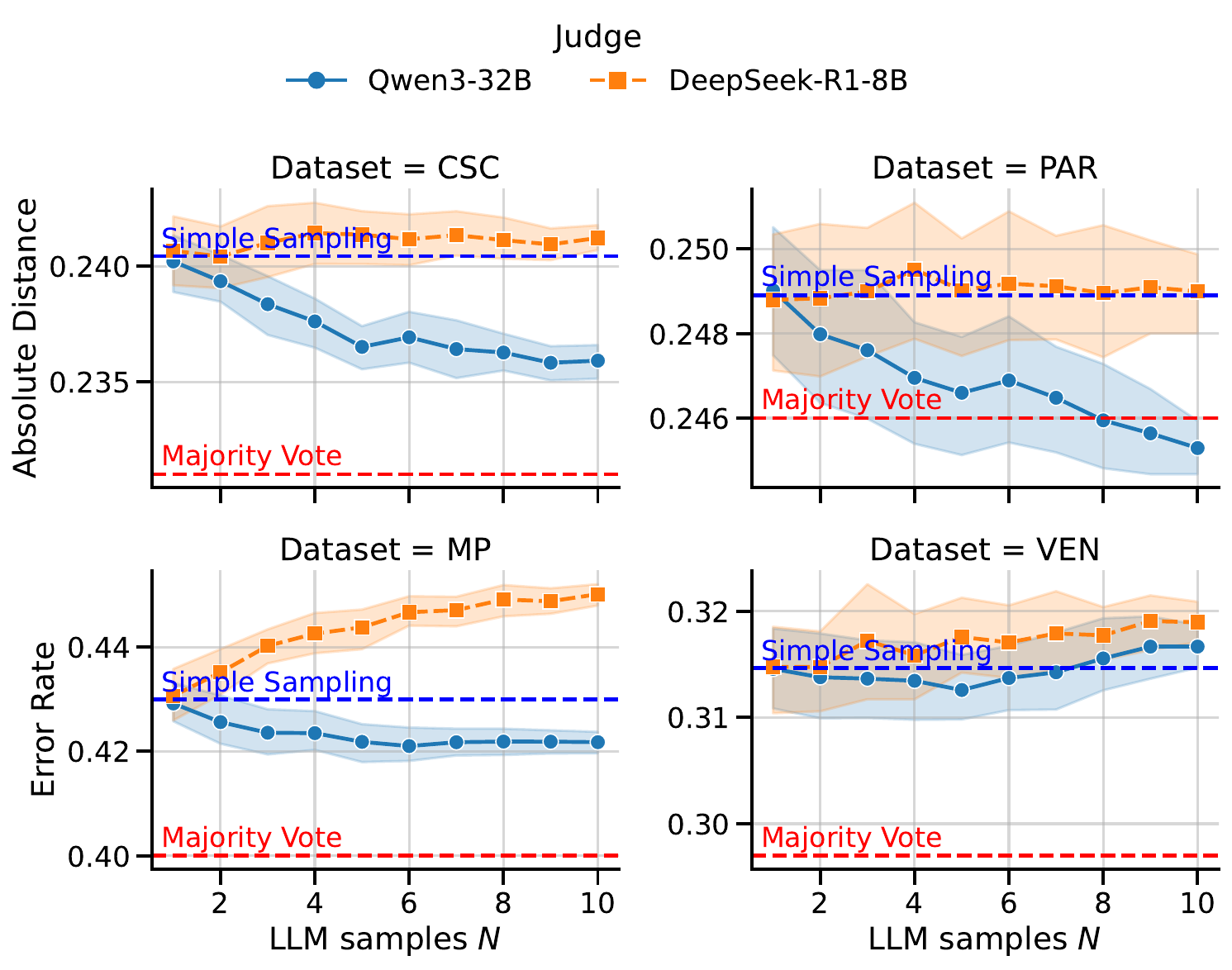}
  \caption{\textbf{Best-of-N Sampling on LeWiDi Tasks}.
  Top: soft-label task. Bottom: perspectivist task.
  Distance metric on the y-axis (lower is better). 
  Higher $N$ should lead to better performance, but does not.
  No consistent pattern emerges across datasets and tasks.
  In red are the test-time scaling benchmarks, which \ac{BoN} generally does not beat.
  The shaded areas show the 0.25 and 0.75 quantiles.}
  \label{fig:bon-samples-vs-perf-nlp}
\end{figure}

The \ac{BoN} sampling method has inconsistent performance in the LeWiDi datasets, as shown in \autoref{fig:bon-samples-vs-perf-nlp}.
\noindent Performance is often flat with the number of samples $N$, or varies wildly with judge model.
\emph{E.g}\ in the MP dataset, the Deepseek judge is consistently \textit{worse} (higher distance) than Simple Sampling on both tasks (soft-label and perspectivist).
\ac{BoN} sampling is only competitive with the benchmarks (red horizontal lines) in a single case (perspectivist task, PAR dataset, Qwen3-32B judge).
These inconsistent results raise the question why step-wise scoring is not effective in the LeWiDi tasks.
For the \ac{BoN} sampling numbers in \autoref{tab:results-test}, we report the Qwen3-32B judge, because it performs slightly better than the Deepseek judge on the perspectivist task.
For the LeWiDi shared task, \textbf{we submitted the predictions for Model Averaging and Majority Voting}, rather than \ac{BoN} sampling.
\subsection{Prediction Diversity}
\begin{figure}[!h]
  \centering
  \includegraphics[width=\columnwidth]{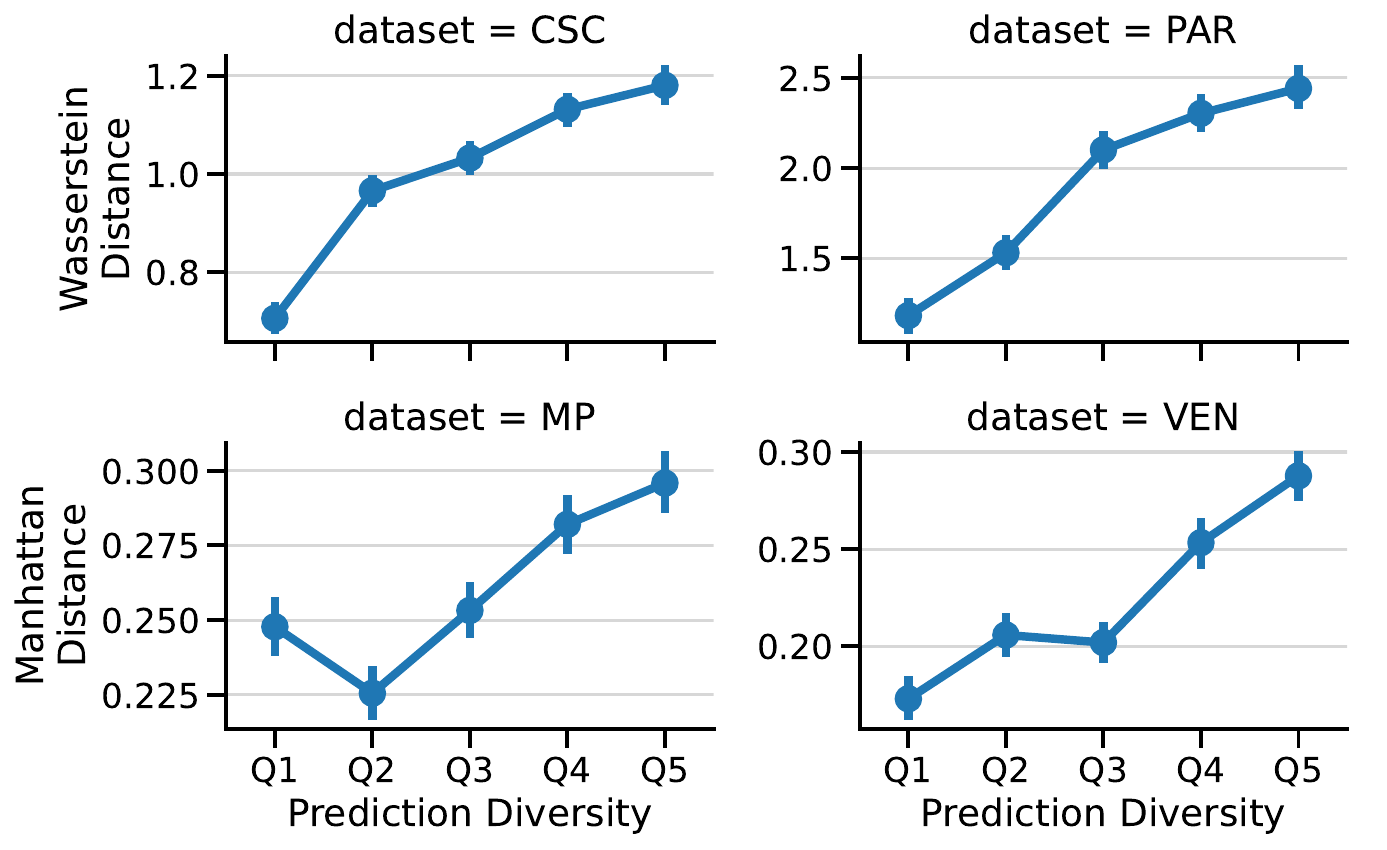}
  \caption{Model performance (lower is better) varies with prediction diversity and is related to the difficulty of the problem.}
  \label{fig:ws-loss-vs-diversity}
\end{figure}

\begin{figure}[!h]
  \centering
  \includegraphics[width=\columnwidth]{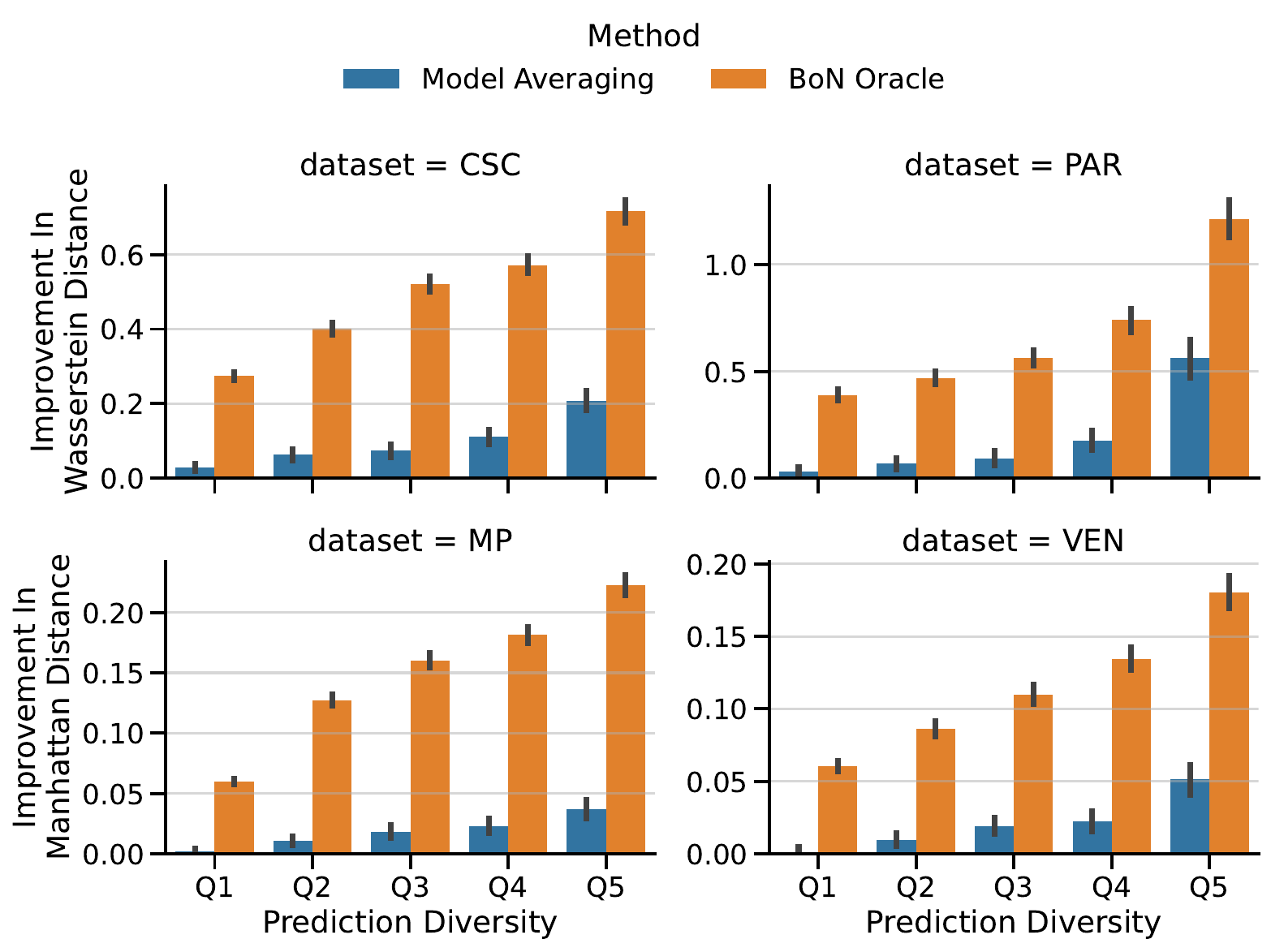}
  \caption{High prediction diversity leads to better performance of test-time scaling methods. Both the upper bound (\ac{BoN} oracle) and Model Averaging benefit from higher prediction diversity. The y-axis shows the improvement over Simple Sampling.}
  \label{fig:improvement-vs-prediction-diversity-quantiles}
\end{figure}

Back to the LeWiDi tasks, we empirically observe that prediction diversity correlates with model performance (\autoref{fig:ws-loss-vs-diversity}):
diversity increases for difficult problems and decreases for easier ones.
For analysis, we binned prediction diversity into five quantiles (but the trends hold for any number of bins).
We document the distribution of prediction diversity across datasets in~\autoref{sec:prediction_diversity}.
Prediction diversity strongly affects test-time scaling methods, as shown in \autoref{fig:improvement-vs-prediction-diversity-quantiles}: 
the \ac{BoN} oracle performance (the upper bound for any \ac{BoN} sampling method) increases with diversity.
The same applies for Model Averaging.

\autoref{tab:model-avg-improvement-vs-prediction-diversity} shows that Model Averaging achieves a significant fraction of the theoretical performance gains dictated by the \ac{BoN} oracle.
For example, in the top quantile of the PAR dataset, Model Averaging achieves 46\% of the performance gains of the \ac{BoN} oracle.


\begin{table}[!h]
\centering
\begin{tabular}{lccccc}
& \multicolumn{5}{c}{Prediction Diversity} \\
\cline{2-6}
Dataset & Q1 & Q2 & Q3 & Q4 & Q5 \\
\hline
CSC & 0.11 & 0.16 & 0.14 & 0.19 & \textbf{0.29} \\
PAR & 0.09 & 0.14 & 0.17 & 0.24 & \textbf{0.46} \\
MP & 0.03 & 0.09 & 0.12 & 0.13 & \textbf{0.17} \\
VEN & 0.03 & 0.11 & 0.18 & 0.18 & \textbf{0.27} \\
\end{tabular}
\caption{Fraction of the \ac{BoN} oracle performance gains that Model Averaging achieves for different datasets and prediction diversities.}
\label{tab:model-avg-improvement-vs-prediction-diversity}
\end{table}

\section{Discussion}
\subsection{\ac{BoN} Sampling Underperformance}
\begin{figure}[h]
  \centering
  \includegraphics[width=\columnwidth]{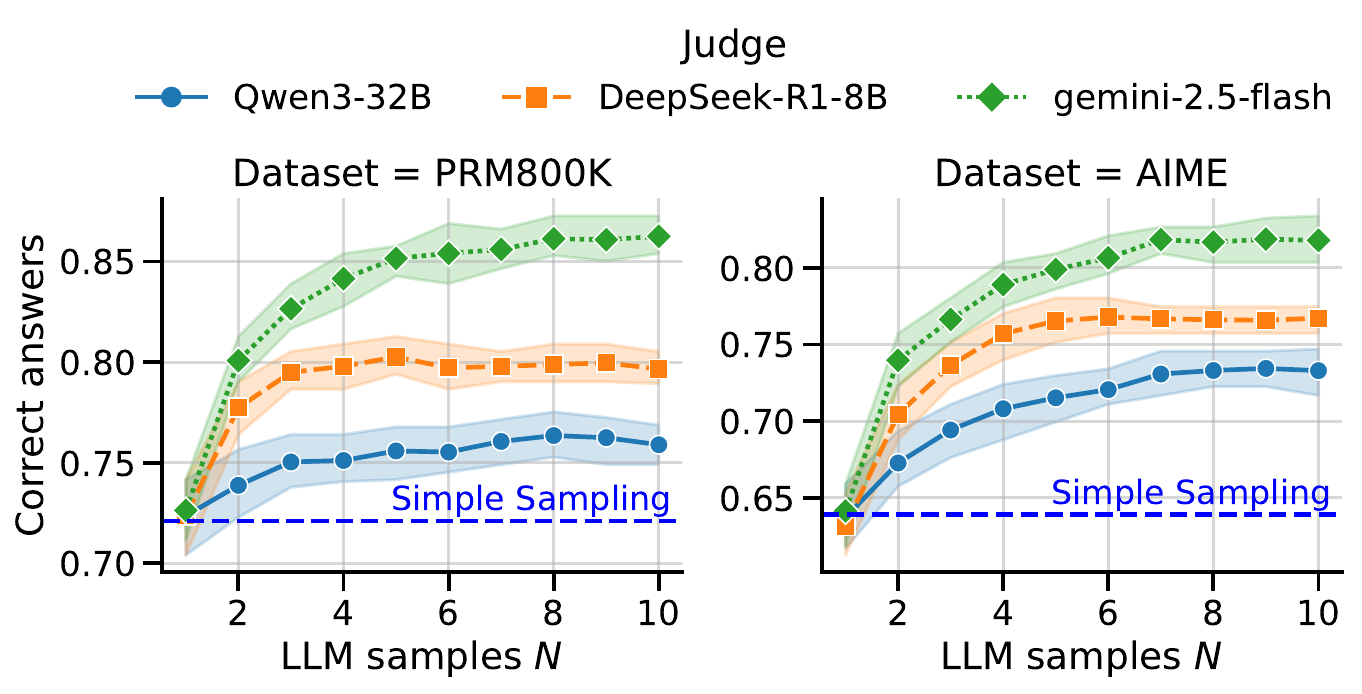}
  \caption{\textbf{Best-of-N Sampling in Mathematics}: The performance of \ac{BoN} sampling (Correct answers, higher is better) improves with the number of samples $N$ and with stronger judges. The shaded area shows the 0.25 and 0.75 quantiles: improvements are consistent.}
  \label{fig:bon-samples-vs-perf-math}
\end{figure}

We were surprised by the underperformance of \ac{BoN} sampling in the LeWiDi datasets.
To verify that we had not made a mistake in our implementation of \ac{BoN} sampling, we ran our method on two math datasets (PRM800K and AIME), as shown in~\autoref{fig:bon-samples-vs-perf-math}.

\noindent The results are in line with~\citet{lightman2023letsverifystepstep}.
Using the best judge and $N=10$ samples, the rate of correct answers jumps by 14\% on PRM800K and by 18\% in AIME.
More information about both datasets is in~\autoref{sec:math-datasets}.

\begin{figure}[!h]
  \centering
  \includegraphics[width=\columnwidth]{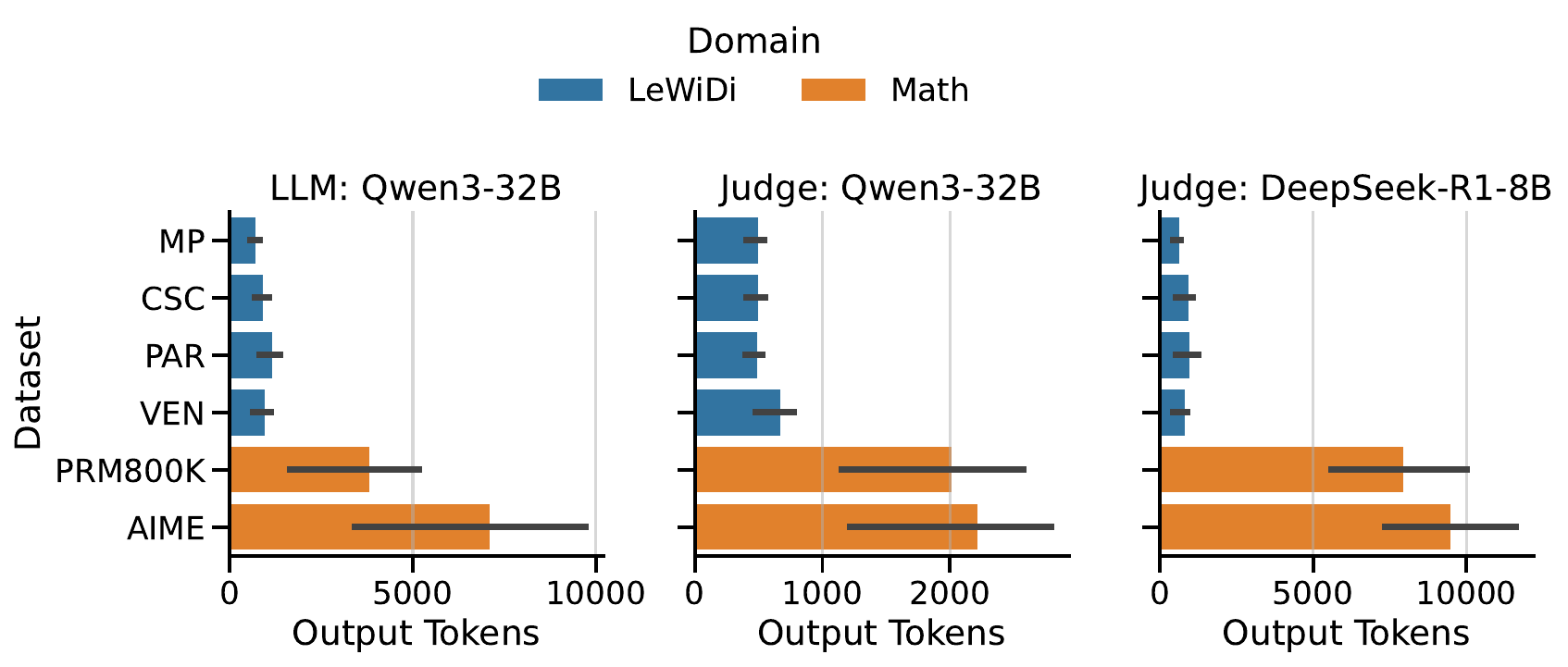}
  \caption{\textbf{Compute Budget} used by \ac{LLM} and judges on different domains. Error bars are the 0.25 and 0.75 quantiles: they show large variability in output length. The models invest an \textbf{order of magnitude} more compute budget into solving AIME problems than in the LeWiDi tasks. Both Qwen3 and Deepseek-R1-8B show this bias.}
  \label{fig:lens-of-responses-and-reasonings}
\end{figure}

Why is \ac{BoN} sampling effective in math, but not in the LeWiDi tasks?
We think the LeWiDi tasks are not inherently harder or more intractable than math problems.
The gap we observe is a failure of \textit{cross-domain generalization}. 
For example, we observed that the shift in domain introduces unexpected side-effects:

\begin{enumerate}
  \item We found qualitative evidence of the \ac{LLM} being \textbf{more vague} in its formulation of \ac{CoT} steps in LeWiDi tasks (see \autoref{sec:vague-cot-steps}).
  When steps are vague, it is harder for a judge to discriminate between good and bad steps.
  During \textit{post-training}, the Qwen3 model was likely never rewarded for summarizing precise arguments around interpretative variation and different perspectives.
  In contrast, we know that Qwen3 has been post-trained to reason on \textit{``[...] math, code, logical reasoning, and general STEM problems.''}~\citep{yang2025qwen3technicalreport}. We find same in the technical reports for Deepseek R1 and Gemini-2.5~\citep{deepseekai2025deepseekr1incentivizingreasoningcapability,comanici2025gemini25pushingfrontier}.
  \item We empirically observe that \ac{LLM}s and judges both spend a higher \textbf{compute budget} (\emph{i.e.}\ they produce more tokens) on reasoning when solving the mathematical tasks than on the LeWiDi tasks
  as shown in \autoref{fig:lens-of-responses-and-reasonings}.
  Since reasoning capabilities are learned during post-training, we hypothesize that this difference is also caused by the standard post-training recipe.
\end{enumerate}

\subsection{Logical Steps in LeWiDi Tasks}
One might argue that step-wise scoring requires a clear boundary between correct and incorrect steps, which is lacking in tasks with strong interpretative variation.
We argue against this for two reasons:  
\begin{enumerate}
    \item Mathematical problem solving is also not always clear-cut. \citet{lightman2023letsverifystepstep} show many steps add no insight or progress, leading them to use an ``okay'' label alongside ``great'' and ``bad''.  
    \item LeWiDi tasks define correctness precisely (\emph{e.g.}, Wasserstein Distance 0). Steps that are logical, plausible, and advance a prediction are ``great'', while vague or unsound steps are ``bad''.
    Previous perspectivist research has also leveraged explanations to judge the validity of annotations~\cite{weber-genzel-etal-2024-varierr}.
\end{enumerate}  
We see no theoretical conflict between perspectivism and step-wise scoring.
Rather, adjusting the method to incorporate perspectivist principles is an avenue for future work.
For example, using different step labels like ``plausible'', ``implausible'', ``vague'', etc.

\section{Conclusion}
We present a systematic evaluation of three test-time scaling methods on the LeWiDi tasks. Our key findings are:
(1) Our prediction diversity metric correlates with test-time scaling performance and problem difficulty on the LeWiDi soft-label task.
(2) Model Averaging and Majority Voting consistently improve \ac{LLM} performance across the LeWiDi tasks.
(3) \ac{BoN} sampling with step-wise scores does not transfer from the domain of mathematics to the LeWiDi tasks, potentially due to vague reasoning steps and lower reasoning compute used.
We hypothesize that this difference is caused by the post-training recipes of current reasoning \acp{LLM}, which lean towards mathematical and logical reasoning.
The performance on datasets with annotation disagreements could potentially be improved by including similar tasks in the post-training recipe.

\clearpage

\section*{Limitations}
We articulated the limitations of \ac{BoN} sampling with step-wise scores in the LeWiDi tasks.
We do not explore prompt optimization thoroughly, because we think that methods should be robust over different prompts.
In terms of the prediction diversity metric, we suggest that authors evaluate the correlation with problem difficulty on their own datasets, since we showed an empirical rather than theoretical relationship. 

\section*{Acknowledgements}
This research is funded by the Bavarian Research Institute for Digital Transformation (bidt).

\bibliography{anthology-1,anthology-2,custom}

\clearpage
\appendix

\section{Sampling Parameters}
\label{sec:sampling-parameters}
\citet{yang2025qwen3technicalreport} suggest two different parameter configurations for Qwen3: for thinking and non-thinking modes.
In early expriments we found almost no difference in performance between both configurations, but observed less variation with the non-thinking configuration, so we used thoese parameters in our experiments: top-k=20, top-p=0.8, temperature=0.7, presence-penalty=1.5.
We used the same parameters for the Deepseek-R1-8B model.
For Gemini-2.5-flash we used the default parameters documented in Google's documentation: top-k=64, top-p=0.95, temperature=1.0\footnote{https://cloud.google.com/vertex-ai/generative-ai/docs/models/gemini/2-5-pro}.

\section{Splitting the \ac{CoT} into Steps}
\label{sec:splitting-cot-into-steps}

To score each step of a \ac{CoT} for correctness, it must be first split into steps.
We instruct the model to answer using a structured format (JSON) with separate fields for the prediction and the \ac{CoT} steps, as shown in \autoref{lst:pred-output-format}.
\promptlistingsmall{listings/pred_output_format.txt}{Output format for the \ac{LLM}}{lst:pred-output-format}
\noindent We found that this approach to get logical steps is more robust than two alternatives: (1) Using \textit{string matching} (\emph{e.g.}\ on double line breaks) to split a \ac{CoT} into steps, because it produces overly granular, incoherent steps, where \emph{e.g.}\ a bulleted list becomes a step on its own.
(2) Using a separate \ac{LLM} to \textit{reformat} the \ac{CoT} into logical steps, because the reformatting model sometimes rephrases and truncates the original \ac{CoT} instead of only reformatting it.
We think that the original \ac{LLM} is best positioned to split its own reasoning process into coherent, logical steps.

One might argue these steps are \textit{constructed ex-post} and do not reflect the model's true reasoning.
However, during a math exam, students are allowed to sketch ungraded work on separate sheets, and then present a clean step-by-step solution.
We follow this same principle, and our math experiments show that the ex-post steps are expressive enough to discriminate good and bad reasoning.

\section{Mathematical Datasets}
\label{sec:math-datasets}
Our method for providing step-wise scores is \textit{completely automated} and requires no human annotations for the \ac{CoT} steps at all.
As \textbf{a sanity check} that our test-time scaling implementation is correct, we also include in our \ac{BoN} evaluation two datasets with mathematical problems, where we expect step-wise scoring to perform very well.
The datasets are: (1) High-school math problems and solutions compiled in the \textbf{PRM800K} dataset by~\citet{lightman2023letsverifystepstep}. 
The problems are originally from the MATH dataset by~\citet{hendrycks2021measuringmathematicalproblemsolving}.
(2) Mathematical problems given to the top 2.5\% to 5\% of high-school students in the US from the American Invitational Mathematics Examination \textbf{(AIME)} compiled by~\citet{aime_1983_2024} and ranging from 1983 to 2024.
The AIME problems are generally more difficult than those in PRM800K.
Many math problems are solved by Qwen3-32B in 10/10 samples, which makes \ac{BoN} sampling unnecessary.
We skipped these problems in our \ac{BoN} evaluation, which is why the horizontal line for Simple Sampling is relatively low in \autoref{fig:bon-samples-vs-perf-math}.

\section{LLM Compliance}
When using an \ac{LLM} with structured output, we need to measure its adherence to the output format of the prompt. We call this \textbf{compliance}.
As we show in \autoref{tab:compliance}, the compliance level varies by dataset and by task.

\begin{table}[h]
\centering
\begin{tabular}{lccc}
\hline
Dataset & Perspectivist & Soft-label \\
\hline
MP & 100.0 & 100.0 \\
CSC & 100.0 & 99.3 \\
VEN & 100.0 & 93.9 \\
PAR & 100.0 & 86.2 \\
\hline
\end{tabular}
\caption{Percentage of compliant predictions sorted by dataset from highest to lowest. The PAR soft-label task is difficult because the weight of 11 classes must sum to 1.0.}
\label{tab:compliance}
\end{table}

We observe near-perfect compliance for the CSC and MP datasets.
The VEN dataset has lower compliance because of the nested strucure of the predictions (one for each \ac{NLI} category).
The lowest compliance is in the PAR datasets, which has 11 classes (-5 to 5, including 0).
We found that the \ac{LLM} outputs correct JSON for PAR, but often the soft-labels did not sum exactly to 1.
We experimented with enforcing \textbf{structured outputs} in vLLM, but initial experiments showed that the \ac{LLM} would sometimes output infinite newline characters until it reached the output token limit, which is valid JSON, so we dropped this constraint.

\section{Model Averaging and Entropy}
\label{sec:entropy_analysis}
Model Averaging has an \textbf{adaptive} flattening effect on the model's soft-labels:
When the model identified a consensus interpretation (regime of low prediction diversity), Model Averaging keeps soft-labels intact (\emph{e.g.}\ peaky).
And when the model's answers are diverse, Model Averaging flattens the soft-labels, which has a hedging effect.
We compare Model Averaging with a naive \textbf{smoothing} method, which flattens a soft-label by averaging it with the uniform distribution, therefore increasing the entropy of the soft-label.

\begin{figure}[!h]
  \centering
  \includegraphics[width=\columnwidth]{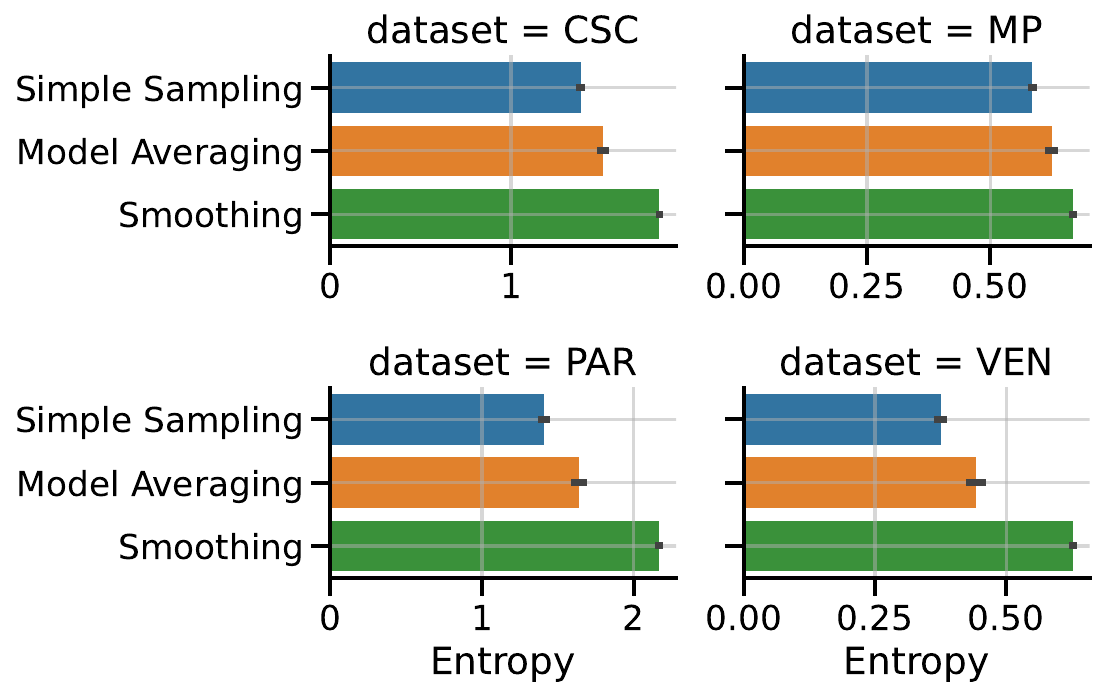}
  \caption{\textbf{Entropy of soft-labels}: We observe that both smoothing (green) and Model Averaging (orange) increase the entropy of the soft-labels, but only Model Averaging improves the model performance.}
  \label{fig:entropy_vs_model_size}
\end{figure}

\autoref{fig:entropy_vs_model_size} shows the entropy of the soft-labels for Qwen3-32B, for different datasets and sampling methods. 
It shows that smoothing the soft-labels does not automatically improve the model performance and that Model Averaging is much more adaptive than the naive smoothing.

\section{Prediction Diversity}
\label{sec:prediction_diversity}
\begin{figure}[!h]
  \centering
  \includegraphics[width=\columnwidth]{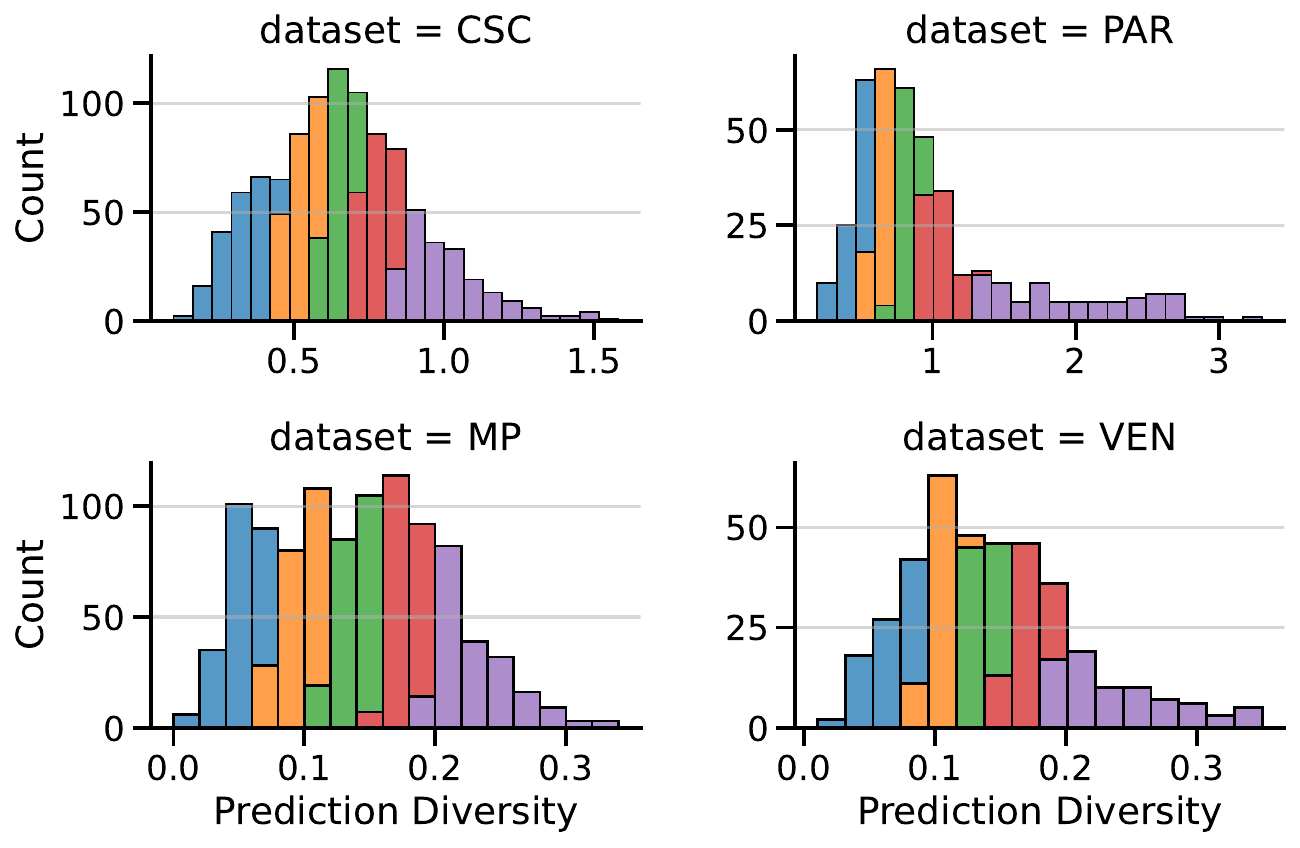}
  \caption{Distribution of prediction diversity by dataset. The distributions follow a normal-like distribution, and the PAR dataset shows a longer tail to the right. The colors indicate the quantiles of the distribution.}
  \label{fig:diversity-distribution}
\end{figure}

The distribution of prediction diversity by dataset is shown in \autoref{fig:diversity-distribution}.
We observe that it is distributed with a single peak in the center, and sometimes has a right tail.

\section{Compute Infrastructure}
We use the vLLM engine~\cite{kwon2023efficientmemorymanagementlarge} to run the models, because of its high throughput, which help us compute $N$ samples per example in parallel. 
vLLM can also be configured to parse the \ac{CoT} and return them separately from the final answer.
All our experiments are run on a single NVIDIA H100 GPU, except for the Qwen3-32B model, which is run on two GPUs.
We called Gemini-2.5-flash over the Google Cloud API.

\section{Prompt Ablations}
\label{sec:prompt-ablations}
We created two prompt variations that could potentially affect performance for interpretative tasks: (1) One variant provides a \textit{dictionary definition}\footnote{Online dictionary: https://dictionary.cambridge.org/} of sarcasm (for CSC), or irony (for MP). (2) The second variant explicitly instructs the model to consider different \textit{perspectives and interpretations}.
\promptlistingsmall{listings/sarcasm-definition.txt}{Prompt Section Defining Sarcasm}{lst:sarcasm-definition}
\promptlistingsmall{listings/consider-perspectives.txt}{Prompt Section to Consider Perspectives}{lst:consider-perspectives}

We perform an ablation analysis to determine the impact of the two prompt variants: (1) first, we remove the prompt section that defines irony and sarcasm and (2) we remove the prompt section about considering diverse perspectives. 

As shown in \autoref{tab:prompt_ablations} \textbf{for ablation 1}, we observe mixed effects: In the CSC dataset, including the definition of sarcasm improves performance, while in the MP dataset, including the definition of irony decreases performance. 
We compute the 95\% confidence interval of mean performance using the bootstrap method to rule out the possibility that performance differences between prompts are a sampling artifact.

\textbf{For ablation 2}, we observe that prompting the model to consider diverse perspectives improves performance in 3 out of 4 datasets (CSC, MP, VEN).
In the PAR dataset, performance is not affected by the prompt section on diverse perspectives.

\begin{table*}[h]
\centering
\begin{tabular}{lllllllll}
& \multicolumn{8}{c}{Dataset} \\
\cline{2-9}
& \multicolumn{2}{c}{CSC} & \multicolumn{2}{c}{MP} & \multicolumn{2}{c}{PAR} & \multicolumn{2}{c}{VEN} \\
Prompt & low & high & low & high & low & high & low & high \\
\hline
Default & \textbf{1.002} & \textbf{1.013} & 0.256 & 0.258 & 1.838 &  1.920 & \textbf{0.217}  & \textbf{0.227} \\
-Definition (ablation 1) & 1.057 & 1.071 & \textbf{0.246} & \textbf{0.248} & - & - & - & - \\
-Perspectives (ablation 2) & 1.027 & 1.040 & 0.263 & 0.265 & 1.840 & 1.924 & 0.236 & 0.247 \\
\end{tabular}
\caption{Two Prompt Ablations: The numbers are mean performance in terms of Wasserstein Distance (lower is better). 
The column ``low'' and ``high'' are the bounds of the 95\% confidence intervals on the performance computed with the bootstrap method.
Ablation 1 shows that including/excluding the definition of irony / sarcasm has mixed effects.
Ablation 2 shows that not instructing the model to consider diverse perspectives has negative consequences in 3 out of 4 datasets.}
\label{tab:prompt_ablations}
\end{table*}

\section{Vague \ac{CoT} Steps}
\label{sec:vague-cot-steps}
Below is a qualitative comparison of two responses to the same problem by Qwen3-32B. \autoref{lst:problem-description} describes the problem (ID=637 of the CSC dataset on sarcasm detection).
\autoref{lst:vague-logical-steps} shows a response with very vague \ac{CoT} steps, which are not strictly wrong, but are so general that they could be applied to any or all problems.
\autoref{lst:specific-logical-steps} shows a different response with very specific \ac{CoT} steps that directly refer to the problem statement and are easier to judge.

Note that these are the structured logical steps, rather than the raw \ac{CoT}.
The raw \ac{CoT} for the vague answer is in \autoref{lst:vague_llm_reasoning} and the one for the concrete answer is in \autoref{lst:concrete_llm_reasoning}.
Neither of the raw \acs{CoT} are vague, so there is no reason why the model should generate vague logical steps from either of them.
We observed this ``vagueness'' behavior particularly in the CSC dataset.

\promptlistingsmall{examples/css_vagueness/problem.json}{Problem Description (Sarcasm Detection)}{lst:problem-description}

\promptlistingsmall{examples/css_vagueness/llm_steps.txt}{Vague Logical Steps}{lst:vague-logical-steps}

\promptlistingsmall{examples/csc_concrete/llm_steps.txt}{Specific Logical Steps}{lst:specific-logical-steps}

\begin{figure*}[!h]
  \centering
  \promptlisting{examples/css_vagueness/llm_reasoning.txt}{Raw \ac{CoT} for Vague Answer}{lst:vague_llm_reasoning}
\end{figure*}

\begin{figure*}[!h]
  \centering
  \promptlisting{examples/csc_concrete/llm_reasoning.txt}{Raw \ac{CoT} for Concrete Answer}{lst:concrete_llm_reasoning}
\end{figure*}

\section{Prompts}
\label{sec:prompts}

\begin{figure*}[h]
\centering
\promptlisting{listings/CSC_t60.txt}{\ac{LLM} prompt for the CSC dataset (soft-label task)}{lst:csc-soft-label}
\end{figure*}

\begin{figure*}[h]
\centering
\promptlisting{listings/CSC_t63.txt}{\ac{LLM} prompt for the CSC dataset (perspectivist task)}{lst:csc-perspectivist}
\end{figure*}

\begin{figure*}[h]
\centering
\promptlisting{listings/judge-prompt.txt}{Judge prompt for scoring reasoning steps (soft-label task)}{lst:judge-prompt}
\end{figure*}

\end{document}